\documentclass[graybox]{svmult}

\usepackage{mathptmx}       
\usepackage{helvet}         
\usepackage{courier}        
\usepackage{type1cm}        
\usepackage{makeidx}         
\usepackage{graphicx}        
\usepackage{multicol}        
\usepackage[bottom]{footmisc}
\usepackage{url}
\usepackage{multirow}
\usepackage{bm}

\usepackage{hyperref}

\usepackage{newtxtext,newtxmath}

\makeindex             

\begin{document}

\title*{Real-time and Continuous Turn-taking Prediction Using Voice Activity Projection}
\author{Koji Inoue, Bing'er Jiang, Erik Ekstedt, Tatsuya Kawahara and Gabriel Skantze}
\institute{Koji Inoue \at Kyoto University, Japan, \email{inoue.koji.3x@kyoto-u.ac.jp}
\and Bing'er Jiang \at KTH Royal Institute of Technology, Sweden, \email{binger@kth.se}
\and Erik Ekstedt \at KTH Royal Institute of Technology, Sweden \email{erikekst@kth.se}
\and Tatsuya Kawahara \at Kyoto University, Japan, \email{kawahara@i.kyoto-u.ac.jp}
\and Gabriel Skantze \at KTH Royal Institute of Technology, Sweden \email{skantze@kth.se}
\\
\\
This paper has been accepted for presentation at International Workshop on Spoken Dialogue Systems Technology 2024 (IWSDS 2024) and represents the author's version of the work.
}
%
%
\maketitle

\abstract*{
A demonstration of a real-time and continuous turn-taking prediction system is presented.
The system is based on a voice activity projection (VAP) model, which directly maps dialogue stereo audio to future voice activities.
The VAP model includes contrastive predictive coding (CPC) and self-attention transformers, followed by a cross-attention transformer.
We examine the effect of the input context audio length and demonstrate that the proposed system can operate in real-time with CPU settings, with minimal performance degradation.
}

\abstract{
A demonstration of a real-time and continuous turn-taking prediction system is presented.
The system is based on a voice activity projection (VAP) model, which directly maps dialogue stereo audio to future voice activities.
The VAP model includes contrastive predictive coding (CPC) and self-attention transformers, followed by a cross-attention transformer.
We examine the effect of the input context audio length and demonstrate that the proposed system can operate in real-time with CPU settings, with minimal performance degradation.
}

\section{Introduction}

Turn-taking is a crucial aspect of human spoken interaction and therefore an important function to model in spoken dialogue systems (SDSs)~\cite{skantze2021review}.
The problem of turn-taking in SDSs involves predicting the end of the user's turn and smoothly initiating the system's turn.
Despite recent progress in large language models (LLMs) that enable the generation of sophisticated responses in SDSs, turn-taking is still typically handled in a simplistic manner.
In practical SDSs, turn-taking is commonly implemented using a simple silence timeout threshold, usually around 1 or 2 seconds, to indicate the end of a turn.
However, silence is not a reliable indicator, as pauses within turns are usually longer than pauses between turns in human-human interaction~\cite{heldner2010pauses,Levinson2015TurnTaking}.
As a result, SDSs often suffer from long response delays or frequent interruptions during pauses.

To address this problem, many proposals have been made for end-of-turn prediction models that consider verbal and non-verbal cues (such as linguistic and prosodic features) of preceding user utterances, in order to predict whether the user is just pausing (a \textit{hold}), or whether the turn is yielded (a \textit{shift}).
The models used for this prediction range from recurrent neural networks (RNNs)~\cite{skantze2017sigdial,masumura2017} to transformers~\cite{ekstedt2020turngpt,sakuma2023slt,muromachi2023interspeech,kurata23_interspeech}.
However, in general, the problem setting mostly involves binary prediction of whether the user's turn ends, which is done at the utterance level.
When implementing this in a spoken dialogue system, it is necessary to determine the appropriate waiting time when the system takes a turn~\cite{raux2012,lala2017sigdial,lala2018icmi,sakuma2023slt}.
Therefore, it is preferable to always perform turn-taking prediction in continuous time frames, rather than at the utterance level.

We have proposed a continuous turn-taking prediction model called voice activity projection (VAP)~\cite{erik2022vap}.
The VAP model utilizes multi-layer transformers to predict the near future voice activities of dialogue participants by processing the raw audio signals from the two speakers in a dyadic dialogue.
However, the processing time of the transformer depends on the length of the input context.
It is unclear whether the VAP model can function effectively in real-time environments for SDSs.
In this demonstration, we showcase the real-time processing of VAP for SDSs in CPU environments, investigating the impact of the input context audio length on performance, ensuring minimal degradation.

\section{Voice activity projection} \label{sec:proposed}

As stated above, the main objective for the VAP model is to predict future voice activity of two speakers in a dialogue, based on raw audio input.
A detailed explanation of the VAP model can be found in its original paper~\cite{ekstedt2023predictive}.
The source codes for the VAP model are publicly available\footnote{\url{https://github.com/ErikEkstedt/VoiceActivityProjection}}.

\begin{figure}[t]
\centering
\includegraphics[width=80mm]{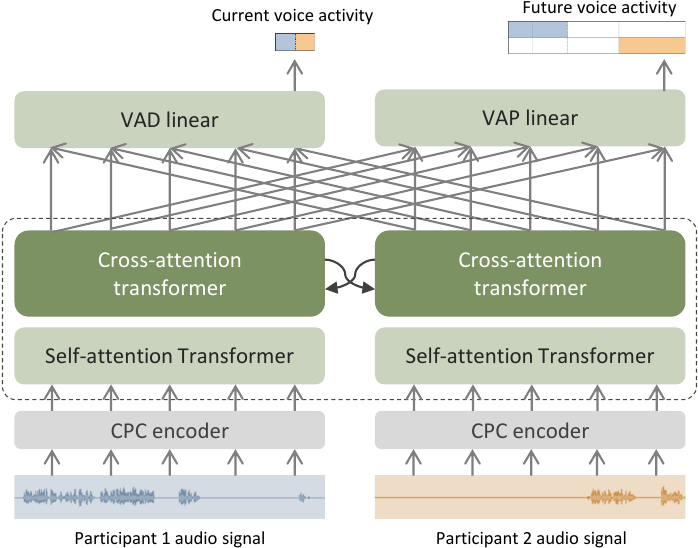}
\caption{Architecture of the VAP model}
\label{fig:architecture} 
\end{figure}

Fig.~\ref{fig:architecture} illustrates the VAP model architecture including a pre-trained contrastive predictive coding (CPC) model for encoding the input audio signal~\cite{riviere2020unsupervised}, followed by separate one-layer self-attention transformer for each channel.
The outputs from the two channels are further processed by a cross-attention transformer, which captures interactive information between the channels~\cite{nguyen2023generative}.
The final output is obtained by concatenating the outputs of the two transformers and passed through linear layers for multitask learning, including the VAP objective and voice activity detection (VAD) subtask.

\begin{figure}[t]
\centering
\includegraphics[width=100mm]{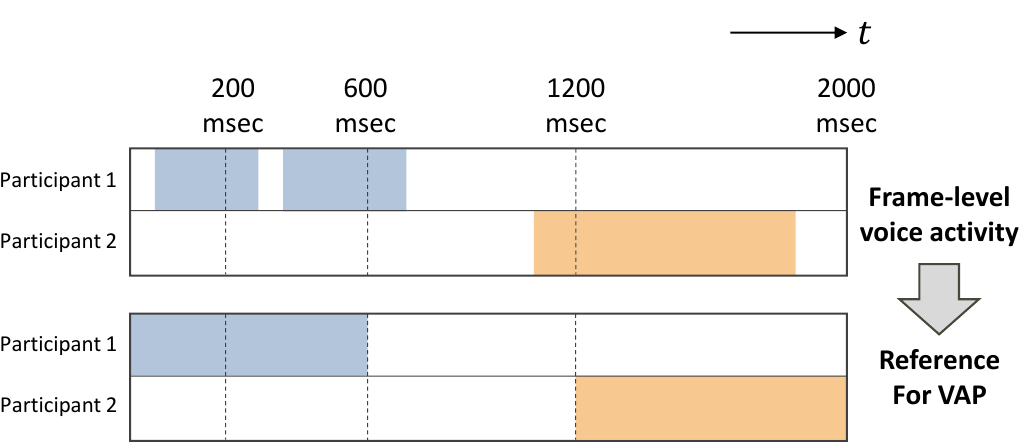}
\caption{Discretizing bins for the VAP model}
\label{fig:state} 
\end{figure}

The VAP model predicts voice activities for two speakers within a two-second time window by predicting the joint activity of both speakers over four binary bins. The objective is to classify the future two seconds into one of 256 possible activations. The bins are discretized as `voiced' or `unvoiced' based on the ratio of voiced frames, as depicted in Fig.~\ref{fig:state}.


The probability distribution over the possible VAP states predicts the turn-taking dynamics. However, it is complex to use and interpret directly.
To simplify, we can sum up the probability values of each participant's bins in the 0-200 msec and 200-600 msec regions.
Then, apply softmax to both sums to obtain $p_{now}(s)$.
This represents a short-term future voice activity prediction for participant $s$ (i.e., ``how likely is the participant to speak in the next 600 msec'').
Similarly, for the 600-1200 msec and 1200-2000 msec bins, we use $p_{future}(s)$ as a slightly longer-term future voice activity prediction.
Note that this is just one example of how the VAP output can be used.

\begin{figure}[t]
\centering
\includegraphics[width=\columnwidth]{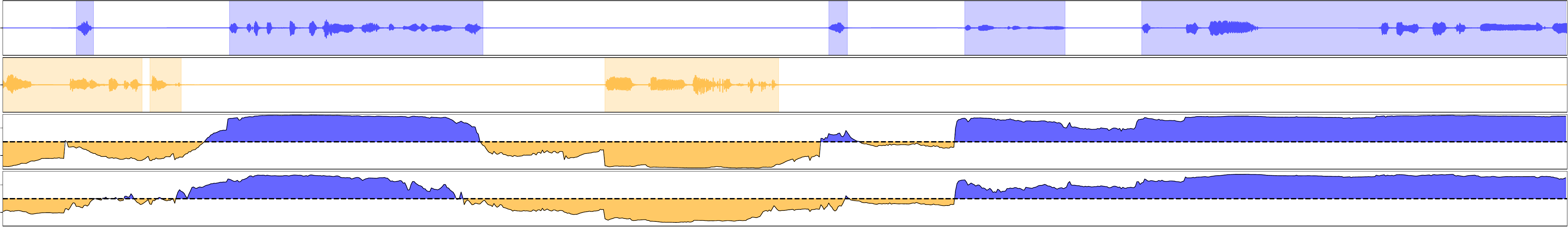}
\caption{Output example of multilingual VAP - Each graph consists of, from top to bottom, input waveforms of both participants, near future voiced probability ($p_{now}$), and future voiced probability ($p_{future}$) among participants.}
\label{fig:output} 
\end{figure}

In Figure~\ref{fig:output}, an example of a GUI for output based on the VAP model is shown. This example depicts the alternating transition of speaking turn from a yellow participant to a blue participant.
At each time frame, the values of $p_{now}(s)$ and $p_{future}(s)$ are indicated.
During the initial transition, the value of $p_{future}(s)$ for the blue participant becomes higher just before the end of the yellow participant's turn.
Immediately after the end of the speech, $p_{now}(s)$ also predicts the transition of turns.
Similarly, the subsequent transition from blue to yellow can be successfully predicted.
There is also a longer mutual pause in the transition from yellow to blue, indicating an ambiguous prediction where $p_{now}(s)$ and $p_{future}(s)$ are kept around even among the two participants.
This suggests that the next speaker is not clear in this case, so it can be said that the model correctly predicts this kind of ambiguous place.
In the subsequent scene where the blue participant continues to hold the speaking turn, even with pauses, the values related to blue remain high, correctly predicting the turn-holding.
In this demonstration, a GUI application is presented to display input waveforms and output model values in real-time.

\section{Performance}

We conducted an investigation into the accuracy and processing speed of the VAP model.
The VAP model utilizes transformers, which leads to slower inference speed as the input data length increases.
In order to mitigate this issue, we performed a study on the accuracy and inference speed of turn-taking prediction by reducing the input sequence length of the transformer.
It is worth noting that a gated recurrent unit (GRU) inside the audio encoder CPC is an auto-regressive model, so we inputted all the audio sequences to the GRU, for durations of up to 20 seconds.

The model in this experiment was trained using the Japanese Travel Agency Task Dialogue dataset~\cite{inaba2023travel}.
The training data consisted of 92.5 hours, with 11.5 hours allocated for both the validation and test sets.
The evaluation scheme is based on previous research~\cite{erik2022vap}, which focuses on predicting the next speaker during periods of mutual silence lasting longer than 0.25 seconds.
The model predicts whether there will be a turn transition or holding by calculating and averaging the value of $p_{now}(s)$ over time during mutual silence.
The higher value between the two participants is then considered the prediction result.
In the test data used for this study, there were 1,023 instances of turn transition and 1,371 instances of turn hold.
The evaluation metric used is balanced accuracy which calculates the accuracies for both positive and negative examples and then takes the average of them.
Note that the random prediction score of this metric would be 0.5.
The VAP model parameters consist of a self-attention transformer with 1 layer for each channel, and a cross-channel transformer with 3 layers.
Both have 4 attention heads and a unit size of 256.
The inference was performed on an Intel Xeon Gold 6128 CPU with a clock speed of 3.40 GHz.

Table~\ref{table:performance} demonstrates the changes in prediction performance and inference time.
Even when the input sequence length is limited to approximately 1 second in the transformer, there is no decrease in prediction performance.
This indicates that the trained VAP model, assuming 50 frames per second in the input, can process in real-time with sufficient inference time.
This result also suggests that the GRU of CPC has likely acquired a certain level of representation regarding the information in the input sequence.
It is important to note that when the input sequence is shortened to less than 0.3 seconds, the prediction performance starts to degrade.
In summary, these results indicate that by restricting the input sequence to around 1 second in the transformer, real-time processing becomes feasible without compromising accuracy.

Furthermore, models for English and Mandarin Chinese have also been developed, making this demonstration multi-lingual.
The English model was trained using the Switchboard dataset~\cite{swb}, while the Mandarin Chinese model was trained using the HKUST Mandarin telephone speech corpus~\cite{hkust}.
Both models yielded similar results as mentioned above.

\begin{table}[t]
\caption{Performance of turn-taking prediction with different input context length}
\setlength{\tabcolsep}{3mm}
\label{table:performance}
\begin{tabular}{ccc}
\hline
\multirow{2}{*}{Input length [sec.]} & \multirow{2}{*}{Balanced accuracy [\%]} & Inference time / frame [msec]\\
& & (real-time factor)  \\
\hline
20.0              & 74.20     &  273.84~~(13.69) \\
10.0              & 75.73     &  \phantom{0}94.93~~(\phantom{0}4.75) \\
\phantom{0}5.0               & 75.01     &  \phantom{0}33.66~~(\phantom{0}1.68) \\
\phantom{0}3.0               & 75.75     &  \phantom{0}30.54~~(\phantom{0}1.53) \\
\phantom{0}1.0               & 76.16     &  \phantom{0}14.61~~(\phantom{0}0.73) \\
\phantom{0}0.5             & 75.41     &  \phantom{0}13.11~~(\phantom{0}0.66) \\
\phantom{0}0.3             & 71.50     &  \phantom{0}12.19~~(\phantom{0}0.61) \\
\phantom{0}0.1             & 62.81     &  \phantom{0}12.45~~(\phantom{0}0.62) \\
\hline
\end{tabular}
\end{table}

\section{Conclusion}

We presented a real-time demonstration of a continuous turn-taking prediction model called voice activity projection.
Conducting investigations on the impact of the sequence length of input to the transformer, we discovered that even with a 1-second input sequence, the prediction accuracy remains unaffected. This allows the model to operate in real-time on a CPU environment.
In the future, we plan to integrate this system into spoken dialogue systems and evaluate the effectiveness of this turn-taking prediction model through dialogue experiments.

\section*{Acknowledgement}

This work was supported by JST ACT-X (JPMJAX2103), JST Moonshot R\&D (JPMJPS2011), and JSPS KAKENHI (JP19H05691 and JP23K16901).

\bibliographystyle{unsrt}
\bibliography{IWSDS2024_KI_DEMO}

\end{document}